**Generation of Accurate Translational Motion for Testing Inertial Sensors**

Swavik Spiewak[1], Stephen Ludwick[2], Glenn Hauer[3], Jakub Mozaryn[4]

[1]*University of Calgary, Canada* ;   [2]*Aerotech, Inc., USA*
[3]*ARAM Systems Ltd., Canada* ;   [4]*Warsaw University of Technology, Poland*

*sspiewak@ucalgary.ca*

**Abstract**

An experimental setup is presented, developed for comprehensive evaluation of high performance inertial sensors for translational motion, accelerometers and geophones. It employs a precision, robust air bearing stage driven by integral brushless DC motor. Experimental results illustrate the performance and capabilities of the setup.

**1    Introduction**

Steadily improving performance of inertial sensors (accelerometers, geophones and gyroscopes) significantly broadens the range of their applications. It also necessitates significant enhancement of the evaluation methods and equipment used throughout the sensor development process. This research is concerned with evaluating accelerometers and geophones used for accurate motion tracking and geophysical sensing.

The sensors of interest feature the frequency bandwidth from DC to several kilohertz and very low nonlinear distortions, down to 0.0001%. Some of them are equipped with 24 bit analog-to-digital converters, which define their best case resolution as $6 \cdot 10^{-2}$ parts per million (ppm). Sensor weight ranges from a few to over 500 grams. Such specifications define a challenging performance envelope for generating motion required for sensor evaluation.  For example, achieving a typical full-scale acceleration (20 m/s$^2$) in 1 Hz sinusoidal motion necessitates using a 507 mm amplitude stroke. At 5 kHz the same acceleration is achieved with just 20.3 nm.  Generating and measuring such diverse displacements with nonlinear distortions below 0.0001% is extremely difficult, in particular when testing heavy sensors. For reference, accurate capacitive sensors for measuring nanometer range displacements feature nonlinear distortions down to 0.05%.

## 2      Experimental setup

There is a broad spectrum of devices that can potentially be employed for evaluating translational inertial sensors. However, most of them do not meet the requirement of low motion distortions. One of the devices successfully used in the past to test high performance digital accelerometers is the ELECTRO-SEIS® shaker (APS Dynamics) shown in Figure 1a [1, 2]. It comprises two modules, an electrodynamic force generator and a passive air bearing stage, which are coupled with a rod. The shaker is a voltage controlled, open loop system without any integral motion sensor. Its architecture with distinctive mechanical modules increases the likelihood of complex broadband dynamics and contamination of signals from the tested sensors.

By comparing various means for generating accurate translational motion we conclude that precision, robust air bearing stages driven by integral brushless DC motors and equipped with accurate displacement sensors are most suitable to satisfy, or at least to reconcile the often conflicting requirements of the tests under consideration. For developing a prototype setup we focus on the 100 mm stroke Aerotech air bearing stage model ABL2010 equipped with Heidenhain optical scale. The stage, shown in Figure 1b, is driven by the NLDRIVE10 digital linear drive.

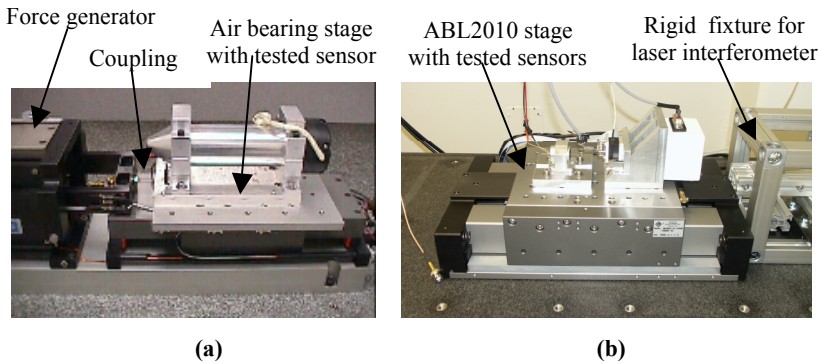

**Figure 1:**  Experimental setups employing APS Dynamics ELECTRO-SEIS® Long Stroke Shaker [1] (a), and Aerotech ABL2010 air stage (b).

To optimize the stage we perform its comprehensive evaluation to detect, characterize, and prioritize various sources of motion errors. This task is greatly simplified and quickened by the A3200 control software (Aerotech) that features a "toolbox" of interactive evaluation and tuning subroutines. We find out for example that residual drag in the guides of compressed air tubing, combined with a conservative tuning of the "off-the-shelf" ABL2010 stage (ample stability margin for operation in diverse conditions) can lead to a mild stick-slip phenomenon such as shown in Figure 2a.

## 3    Representative results

To maximize the accuracy and reliability of the measurements the developed system employs two independent sensors of stage motion. One of them is an integral optical scale of the stage (Heidenhain) set to the resolution 2.5 nm per pulse. This scale is not co-linear with the measured displacement, which can cause the Abbe's errors. The other sensor is a fiber-optic, differential laser Doppler vibrometer (Polytec) mounted co-linearly with the measured displacement. Readouts of both sensors are nearly identical (overlap on the plots). They are shown in Figure 2 for the case: (a) before any stage tuning and, (b) after a reduction of the drag caused by compressed air tubing and cables, and optimization of the standard control algorithm.

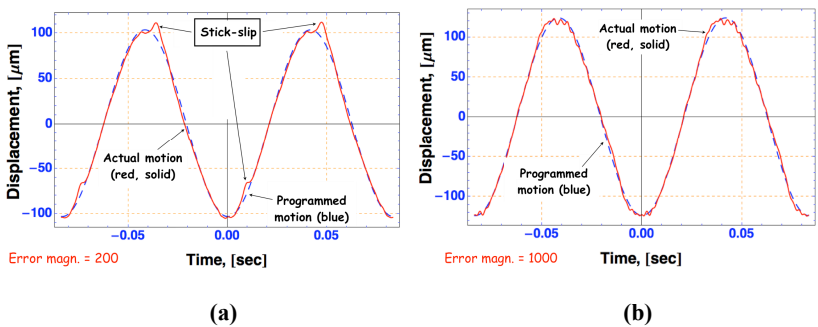

(a)                                         (b)

**Figure 2:** Example plots of the measured displacement before stage tuning (a), and after its basic optimization (b). Deviations of the measured signals from the nominal waveform are magnified 200 and 1000 times in the former and latter case, respectively.

The nominal stage displacement is in both cases a 12 Hz sine wave with 237 µm peak-to-peak travel, represented by the blue dashed line. Such signal is typically used to evaluate geophysical inertial sensors. Two periods of the waveform are plotted. The actual displacement measured by the optical scale and laser are shown with the error magnified 200 times (case a) and 1000 times (case b). Without such magnification the actual and nominal displacements can not be distinguished.

Average deviations between the nominal and actual motion of the optimized stage, which are shown in Figure 3a for two periods of the sinusoidal excitation, show strong periodicity at the harmonics of the excitation. The form of these deviations indicates an absence of stick-slip in the system. The deviations can be further decreased to a few nanometers peak-to-peak by fine tuning of the stage controller at the specific frequency of the excitation signal, 12 Hz in this case. However, a more robust and effective approach is also available. It is Iterative Learning Control (ILC) implemented through the Motion Designer package (Aerotech).

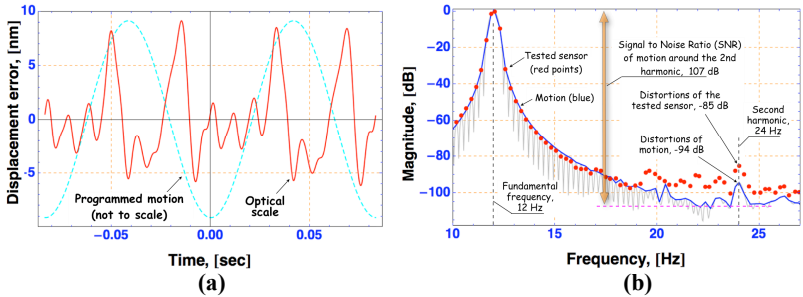

**Figure 3:** Deviations of the stage position from the nominal waveform (a) and representative frequency spectra of the investigated signals (b).

### 4    Conclusions

Robust mechanical integration of the guidance, actuation and measurement functions in the ABL2010 air stage facilitates achieving low distortions of motion that are needed for the evaluation of high performance inertial sensors. The prototype system features at present the signal-to-noise ratio over 100 dB for testing velocity sensors (110 dB for displacement) and motion distortions at individual harmonics below -94 dB for velocity (-100dB for displacement), as shown in Figure 3b. Such performance assures reliable evaluation of sensors with very low distortions. In the illustrated case the total harmonic distortions of the tested accelerometer (output signal converted to velocity) are only 0.0052%, mainly at the 2nd harmonic. Nonetheless they are clearly distinguished from the distortions of motion. Fine tuning of the stage and incorporation of the Iterative Learning Control hold the promise to extend the range of measurable distortions below -100 dB, or less than 0.001% in the case of velocity distortions (below -110 db and 0.0005% for displacement).

**References:**

[1]  J.C. Gannon et al., "A Seismic Test Facility", SEG 1999 Expanded Abstracts, 1999.
[2]  M.Q. Feng, D-H. Kim, "Novel fiber optic accelerometer system using geometric moir'e fringe", Sensors and Actuators, A 128, 2006, pp. 37–42.